\documentclass[10pt,twoside]{article}

\usepackage{times}
\usepackage[utf8]{inputenc}
\usepackage[T1]{fontenc}
\usepackage{graphicx}

\usepackage{siunitx}
\usepackage{hyperref}
\usepackage{cleveref}
\usepackage{coria-taln2025}
\usepackage[french,english]{babel} %frenchb deprecated, use french instead
\usepackage{comment}
\usepackage{multirow, makecell}

\title{Summarization for Generative Relation Extraction in the Microbiome Domain}
\author{Oumaima El Khettari\up{1}\quad Solen Quiniou\up{1}\quad Samuel Chaffron\up{1}\\
  {\small
    (1) Nantes Université, École Centrale Nantes, CNRS, LS2N, UMR 6004, F-44000 \\ 
    \texttt{
      oumaima.el-khettari@ls2n.fr, solen.quiniou@ls2n.fr, samuel.chaffron@ls2n.fr \\ 
}}}
\newcommand{\secrev}[1]{\textcolor{purple}{[#1]}}

\begin{document}
\maketitle

\resume{
\textbf{Résumé automatique pour l’extraction générative de relations dans le domaine du microbiome}\\
Nous explorons une approche générative pour l’extraction de relations, adaptée à l’étude des interactions dans le microbiote intestinal, un domaine biomédical complexe et faiblement doté en données annotées. Notre méthode s’appuie sur le résumé automatique par des grands modèles de langage, pour affiner le contexte, avant d’extraire les relations via une génération guidée par instruction. Les premiers résultats sur un corpus dédié montrent que le résumé automatique améliore les performances de l’approche générative en réduisant le bruit et en orientant le modèle. Cependant, les méthodes d’extraction de relations à l'aide de modèles de type BERT restent plus performantes. Ce travail en cours met en évidence le potentiel des approches génératives pour l’étude de domaines spécialisés en contexte de faibles ressources.
}

\abstract{}{
\vspace{-0.5cm}
We explore a generative relation extraction (RE) pipeline tailored to the study of interactions in the intestinal microbiome, a complex and low-resource biomedical domain. Our method leverages summarization with large language models (LLMs) to refine context before extracting relations via instruction-tuned generation. Preliminary results on a dedicated corpus show that summarization improves generative RE performance by reducing noise and guiding the model. However, BERT-based RE approaches still outperform generative models. This ongoing work demonstrates the potential of generative methods to support the study of specialized domains in low-resources setting.
}

\motsClefs
  {Extraction de relations générative, Ajustement par instruction, Domaine à faible ressources, Microbiome}
  {Generative Relation Extraction, Instruction-tuning, Low-Resource Domain, Microbiome}

\section{Introduction}

The biomedical field is witnessing a rapid expansion in the study of specialized subdomains under the scope of NLP~\cite{he-etal-2023-medeval}, such as the microbiome interactions~\cite{hogan2024midred}. The study of the gut microbiome is of high importance~\cite{heintz2018human}, as it is a key modulator of human health, influencing gastrointestinal diseases like Inflammatory Bowel Disease, metabolic and neurological conditions such as Parkinson’s disease and depression~\cite{shreiner2015gut,goralczyk2022microbiota,borrego2025human}. Its role in immune regulation, nutrient metabolism, and the gut-brain axis has made it a focal point for biomedical research. However, the microbiome remains insufficiently characterized, especially at the species interaction level, due to the vast diversity of microbes~\cite{zhang2025large}, variation in microbial communities from one individual to another, and the scattered, often implicit, nature of knowledge in scientific publications. %\firstrev{(REMARQUE1)}
The areas involving this subdomain are often underrepresented in large-scale annotated corpora, making them inherently low-resource for tasks such as information extraction. Among these, relation extraction (RE) is essential for structuring biomedical knowledge and supporting downstream applications like knowledge graph construction and hypothesis generation.

%\firstrev{}
Traditional RE methods typically rely on supervised learning with manually annotated data, which is often scarce or unavailable in emerging subdomains. However, the recent rise of LLMs, pre-trained on extensive corpora, has opened new avenues for addressing low-resource scenarios. These models can be adapted to information extraction task to be performed in a generative manner~\cite{zhu2023large,xu2024large}. In a low resource setting, the generative paradigm can offer flexibility in dealing with unseen relation types and complex, context-rich inputs. In this work, we explore how such models can be leveraged to improve RE in a low-resource biomedical domain, particularly through a generative approach using summarization and instruction tuning. %\secrev{}
The contributions of this work are as follows: 
\begin{itemize}
    \item We conduct a comprehensive comparison of zero-shot, instruction-tuned, and encoder-based approaches for document-level biomedical relation extraction, highlighting their strengths and limitations ;
    \item We show that combining summarization with instruction tuning significantly improves generative model performance, especially for smaller models in low-resource settings ;
    \item We provide insights into model behavior, focusing on hallucinations, output consistency, and adherence to predefined labels in generative models.
\end{itemize}

%%================================================================
\section{Related Work}
RE is a fundamental task in biomedical NLP, that consists in identifying and classifying semantic relations between entities of interest. Since this task is modeled as a classification problem, the dominant approaches have used supervised classification models, such as convolutional neural networks~\cite{liu2016drug,gu2017chemical}, recurrent neural networks~\cite{jettakul2019relation,mandya2018combining}, and Transformer-based encoders~\cite{lee2020biobert, gu2021domain, beltagy2019scibert}.

More recently, the emergence of LLMs has shifted the paradigm from discriminative to generative approaches in RE~\cite{wadhwa-etal-2023-revisiting}. For example, \citet{asada2024enhancing} investigated RE, using GPT models, from standard biomedical RE datasets, including EU-ADR~\cite{van2012eu}, GAD~\cite{bravo2015extraction}, and ChemProt~\cite{islamaj2019overview}. While their use shows promise, they underperform compared to domain-specific models like BioBERT~\cite{lee2020biobert} and PubMedBERT~\cite{gu2021domain}. Another study proposed a novel RE method enhanced by LLMs, incorporating techniques such as in-context few-shot learning and chain-of-thought reasoning to improve performance~\cite{zhang2024study}. %\secrev{}
More specifically, the study of protein-protein interactions as been treated as a binary task \cite{rehana2024evaluating,chang_influence_2025}. \citet{rehana2024evaluating} compared Masked Language Models with different 2 GPT models using different techniques, and concludes that BERT models perform better on complex datasets, while GPT models can be just as effective on simpler ones. \citet{chang_influence_2025} investigated the effect of prompt engineering on the same task and showed that domain-specific prompts significantly enhance the performance of two GPT models. However, a key limitation of these generative approaches is that they are usually evaluated on standard biomedical RE datasets that are either binary or with a limited number of relation types. \citet{brokman2025benchmark} applied their method to several biomedical datasets, including BioRED~\cite{luo2022biored}, a manually annotated corpus with eight relation types and four entity types. This makes BioRED one of the datasets most comparable to our use case in terms of entity and relation diversity. On this dataset, reported F1-scores remained below 10\% with GPT-4 and reached 23.2\% with OpenAI o1. It is important to note, however, that their evaluation is end-to-end, including both named entity recognition and RE, rather than focusing only on relation classification. Furthermore, BioRED does not focus on a specific low-resource subfield.
%However, a key limitation of these generative approaches is that they are typically evaluated on standard biomedical RE datasets, which are either binary or limited in the number of relation types. Moreover, these datasets often lack domain specificity, focusing on general biomedical content rather than specialized subfields. This constrains the evaluation of generative models' ability to handle complex relation types in low-resource specialized biomedical domains.

%\subsection{Summarization in the Era of LLMs}
%\secrev{}
LLMs have significantly advanced text summarization, moving beyond extractive methods~\cite{erkan2004lexrank, mihalcea2004textrank, nallapati2017summarunner} to enable fluent and context-aware generation~\cite{lewis2019bart, gupta2019abstractive}. 
Modern summarization systems can synthesize information across sentences, rephrase content, and adapt to various domains, including scientific and medical texts~\cite{goyal2022news, zhang2024benchmarking}. Recent directions further explore personalized~\cite{pu2023summarization}, human-in-the-loop~\cite{chen2023human}, and real-time summarization, reflecting a shift toward more interactive and user-adaptive systems. These advances highlight the role of LLMs in redefining summarization as a dynamic and domain-sensitive task.

%\subsection{Relation Extraction and Summarization}
Summarization and RE share similar challenges, as both require identifying and condensing key information from text~\cite{zhang-etal-2023-enhancing}. LLMs perform well in summarization tasks due to their extensive pretraining on diverse and domain-specific corpora~\cite{van2023clinical, tang2023evaluating, shaib2023summarizing}, but their effectiveness in structured information extraction tasks like RE remains limited~\cite{naguib2024few, ma-etal-2023-large, zhang-etal-2024-unexpected}. The combination of summarization and RE, two tasks with clear conceptual overlap in the generative paradigm, remains largely unexplored, especially in the context of closed generative RE~\cite{li-etal-2023-revisiting-large}.
%To improve RE, recent efforts have focused on refining prompting strategies in few-shot settings.

%%================================================================
\section{Methodology}
We explore two strategies for performing RE using LLMs: a direct approach and a summarization-based approach.

\subsection{Direct RE Approach}
In the direct RE setting, we adopted a zero-shot framework. The model is prompted with an instruction, the input passage, and a list of candidate relation types. Then, it is expected to identify the correct relation between the target entities, without additional training.
We designed a prompt to elicit the relation type between two target entities within a given text passage. The prompt consists of three main components:
\begin{itemize}
    \item Task instruction: Instructs the model to generate the class of the relation between {entity1} and {entity2} based on the input paragraph.
    \item Constraints: Specifies two requirements (i) the model must output only the relation label, and (ii) no justification should be provided. Classes: Increase, Decrease, Stop, Start, Improve, Worsen, Presence, Negative\_correlation, Affects, Causes, Complicates, Experiences, Interacts\_with, Location\_of, Marker/Mechanism, Prevents, Reveals, Treats, Physically\_related\_to, Part\_of, Possible, Associated\_with, None.
    \item Input: A passage excerpt from the source article containing the target entities.
\end{itemize}
%Solen : ce serait bien de donner un exemple réel de prompt (du coup, je ne pense pas qu'une figure soit nécessaire pour illustrer cette approche)

\subsection{Summarization-Based RE Approach}
The summarization-based approach is  a two-step process:
\begin{enumerate}
    \item A summarization step: an LLM is instructed to generate a concise summary that captures the relation between a pair of entities, based on the surrounding context. This step reduces input noise and mitigates hallucination by focusing on the core semantic relation;
    \item A fine-tuning step: a model is instruction-tuned on relation classification task, and is given the generated summaries as input.
\end{enumerate}
This approach relies on a lighter input than the previous one, as the relation types are learned during the fine-tuning step and are not included in the prompt.

%In contrast to the direct approach, where prompts can become lengthy and negatively affect model performance, the summarization-based method aims to minimize prompt size by relying on open-ended instructions. Each step of the process is guided by a tailored prompt.
%
During the summarization step, the model is prompted to generate a concise summary that captures the relation between the two target entities, based on the surrounding context.
In the instruction-tuning step, the model is prompted as if it were a biology expert. It is told that the provided paragraph describes a relation between two entities and is instructed to identify the expressed relation. The input paragraph is provided without any predefined class list, as the model is expected to learn the relation types during training. The prompt is designed as follows: 
\begin{itemize}
    \item System Role: You are a biology expert. You role is to answer the following instruction.
    \item Task instruction: This is a paragraph describing the relation between Entity1 and Entity2. Find the expressed relation in the text.
    \item Input: A passage excerpt from the source article containing the target entities.
\end{itemize}
%Solen : ce serait bien aussi de donner un exemple réel du prompt de chacune des 2 étapes (du coup, je ne pense pas qu'une figure soit nécessaire ici aussi, pour illustrer cette approche)

%%================================================================
\section{Experimental setup}
\subsection{MicrobioRel Corpus}
We conducted our experiments on MicrobioRel\footnote{\url{https://github.com/Stan8/MicrobioRel-dataset}}, a domain-specific corpus for multi-class RE in the microbiome-related biomedical literature. The dataset consists of manually annotated excerpts from scientific articles, covering 22 relation types across six entity categories: species, diseases, chemicals, mutations, genes, and cell lines. With only 1,994 annotated relations in total (see Table~\ref{tab:relcount}, in Appendix~\ref{dataset}), MicrobioRel places this study in a low-resource setting, targeting a biomedical subdomain that remains largely unexplored in NLP research. %The distribution of relation classes is shown in 
%\commenaireSolen{Comme le corpus MicrobioRel n'est pas l'objet du papier, on peut mettre la table 1 en annexe, ce qui fera gagner de la place: fait}

\begin{comment}

\begin{table}[!ht]
\centering
\begin{tabular}{lc|lc}
\hline
\textbf{Relation type} & \textbf{Count} & \textbf{Relation type} & \textbf{Count} \\
\hline
Associated\_with & 210 & Possible & 56 \\
Part\_of & 204 & Worsen & 51 \\
Affects & 202 & Marker-Mechanism & 41 \\
Increase & 185 & Prevents & 39 \\
Location\_of & 146 & Physically\_related\_to & 31 \\
Causes & 144 & Negative\_correlation & 23 \\
Experiences & 132 & Treats & 17 \\
Decrease & 121 & Complicates & 10 \\
Start & 118 & Presence & 9 \\
Improve & 102 & Reveals & 7 \\
Interacts\_with & 86 & Stop & 6 \\
\hline
\end{tabular}
\caption{Relation label counts in the MicrobioRel corpus}
\label{tab:relcount}
\end{table}
\end{comment}
%\firstrev{}
All named entities in MicrobioRel were initially pre-annotated using PubTator~\cite{wei2019pubtator}. As shown in \cite{el-khettari-etal-2023-building}, PubTator2 demonstrates strong performance in recognizing species mentions, which are one of the most important and challenging entity types in our study, alongside diseases. This step is followed by manual correction by domain experts under specific guidelines, to ensure the correctness of entities participating in relations. For relation annotation, annotators were presented with paragraphs containing at least two entities. 
First, annotators were instructed to identify all relevant relations by selecting pairs of entities that could plausibly be linked. Once a relation was established, it was then categorized according to a predefined schema. A "None" class was automatically introduced to support negative examples that could not be annotated manually due to the complexity of identifying all such instances exhaustively. This was done by generating all possible pairs of entities within a paragraph that were not annotated as being related. To avoid overwhelming annotators with negative examples, the number of "None" instances was limited proportionally to the length of each paragraph.% (REMARQUE2)
An example from MicrobioRel is given in Appendix~\ref{example}.

\begin{comment}
Here is an example from MicrobioRel: 
\begin{quote}
\textbf{Text:} "During PD development, microbiota changes can be accompanied by reduced concentrations of branched-chain amino acids (BCAAs) and aromatic amino-acids in comparison with the healthy control group."

\textbf{Entity 1:} PD \\
\textbf{Entity 2:} branched-chain amino acids \\
\textbf{Annotated Relation:} Decrease
\end{quote}
\end{comment}

\subsection{Models}
To prevent biasing the model by presenting only a subset of relation classes in demonstrations, we use a zero-shot framework for direct RE, using Llama2 13B~\cite{touvron2023llamab}, Llama3 8B, Llama 3.2-3B-Instruct ~\cite{dubey2024llama}, Mistral 7B-Instruct~\cite{jiang2023mistral7b}, and BioMistral 7B~\cite{labrak2024biomistral}.
For the initial summarization task, we employ Llama 3.1 in a few-shot setting. The demonstration examples provided follow a simple structure: \textit{Entity1 Relation Entity2}, expressed in a single sentence, to avoid adding constraints on the output structure.
Then, instruction-tuning is performed using parameter-efficient fine-tuning techniques, specifically the Unsloth framework\footnote{\url{https://github.com/unslothai/unsloth}} and LoRA, applied to the Llama-3 3B Instruct, Mistral 7B, and Llama3 8B models, with configurations adapted to each model’s architecture. We choose to instruction-tune smaller models due to computational constraints and the limited size of the MicrobioRel dataset, which makes smaller models more suitable for effective adaptation.
+\secrev{}
Furthermore, we compare these models to biomedical BERT-based models, BioBERT~\cite{lee2020biobert}, SciBERT~\cite{beltagy2019scibert}, PubMedBERT~\cite{gu2021domain} and BioLinkBERT~\cite{yasunaga2022linkbert}, fine-tuned for 30 epochs, using a fixed batch size of 4 and a learning rate set to \num{1e-5}.%}

\subsection{Evaluation metrics}

To assess the quality of the generated summaries, we employ BERTScore~\cite{zhang2019bertscore} and cosine similarity. These metrics respectively capture fine-grained semantic alignment and overall representational similarity between the generated summaries and the original texts. 
We also report the F1-BERTScore and cosine similarity scores prior to RE, in order to evaluate how well the summaries capture relational information. Evaluation of the RE task is carried out using weighted precision, recall, and F1-score.  %\firstrev{
Under an exact match criterion, a prediction is considered correct only if it matches the gold relation label. Although, human evaluation could provide qualitative insights, we focus on strict, reproducible classification performance.
%}
%\[
\begin{equation}
\text{Weighted Metric} = \sum_{i=1}^{n} \left( \frac{\text{support}_i}{\text{total support}} \times \text{metric}_i \right)
\label{eq:metric}
\end{equation}
%\]

%\firstrev{
In the weighted setting, the precision, recall, and F1-score are computed as the weighted averages of the corresponding per-class scores, where the weights are based on the support of each class (see Equation 1). Note that weighted F1 is derived from individual per-class F1 scores, not from weighted precision and recall. %(REMARQUE4)} 
Metrics are computed after post-processing the LLM outputs to remove text beyond the predicted relation class. Only exact matches with the gold labels are considered correct.

%%================================================================
\section{Results and discussions}
\subsection{Relation Extraction Performance}

Table~\ref{results} presents the performance of the proposed RE approaches, comparing the direct prompting method, the summarization-based pipeline, and BERT-based models across the 23 relation classes. 

\begin{table}[b!ht]
    \centering
    \begin{tabular}{llccc} 
        \hline
        \textbf{RE Approach} & \textbf{Models} & \textbf{P} & \textbf{R} & \textbf{F1}\\
        \hline
        \multirow{5}*{Direct Approach} & Llama2 13B & \textbf{39.2} & \textbf{15.2} & \textbf{14.3} \\
            & Llama3 8B & 11.3 & 11.2 & 8.34 \\
            & Llama 3.2-3B-Instruct & 4.9 & 7.47 & 3.78 \\
            & Mistral 7B-Instruct & 26.9 & 10.4 & 4.95 \\
            & BioMistral 7B & 9.41 & 9.87 & 7.07  \\
        \hline
        \multirow{3}*{Summarization-Based} & Llama3 8B & 34.6 &    23.5 & 24.51 \\
        & Llama 3.2-3B-Instruct & 62.3 & \textbf{60.2} & \textbf{59.7} \\
         & Mistral 7B-Instruct & \textbf{62.9} & 46.9 & 44.6 \\
        \hline \hline
        \multirow{4}*{Regular Fine-tuning} 
         & PubMedBERT & \textbf{71.9} & \textbf{71.7} & \textbf{71.3} \\
         & BioLinkBERT & 71.2 & 71.2 & 70.8 \\
         & SciBERT & 69.7 & 69.3 & 69.0 \\
         & BioBERT & 68.1 & 68.5 & 67.8 \\
        \hline
    \end{tabular}
    \caption{Weighted Precision, Recall, and F1-scores (in \%) of RE models on the test set of MicrobioRel. Regular fine-tuning refers to supervised training on the task, while other rows correspond to generative models used with or without instruction tuning.}
    \label{results}
\end{table}

Results highlight the comparative strengths and weaknesses of different approaches to RE in the microbiome domain. Traditional biomedical models such as PubMedBERT, BioBERT, SciBERT, and BioLinkBERT continue to outperform generative models, with PubMedBERT achieving the highest F1 score of 71.3\% . This confirms the effectiveness of domain-specific pre-training and task-aligned fine-tuning for biomedical RE.

\paragraph{Direct Approach: Zero-Shot Limitations\\}

Zero-shot RE yielded in poor results for both versions of Llama. Llama2, in particular, exhibited frequent hallucinations, defined here as generating relation types outside the predefined label set. %\firstrev{
These hallucinations are not limited to under represented relation types, nor do they consistently affect a specific subset of relations. Instead, they depend more on the context in which the relation occurs.
%} 
They can be classified into two categories: (i) relation types that are semantically similar to valid classes (e.g., \textit{Reduce} instead of \textit{Decrease}), and (ii) context-specific phrases that imply a link but do not correspond to any annotated relation. These issues highlight the lack of strict adherence to instructions and class constraints. Moreover, Llama2 predictions were biased toward more general relation types, resulting in higher recall on dominant classes but limited ability to distinguish fine-grained, domain-specific relations. In contrast, Llama3 produced more consistent outputs that were closer to the target label set. This reduced hallucination rate eased post-processing and demonstrated improved alignment with the task. Notably, Llama3 made more balanced use of the available relation classes but sometimes at the expense of accuracy on broader categories. Overall, the performance of Llama models improved with the increase in model size.

In this setting, Mistral 7B-Instruct showed strong bias toward the \textit{associated\_with} class (284 instances predicted), resulting in low label diversity and poor F1 of 4.95\%. BioMistral 7B produced more varied outputs, favoring \textit{increase} (240 instances), but it also struggled with precision and class balance since it generated only six relation types out of 23. Overall, both models underperformed, highlighting the difficulty of zero-shot generative RE in specialized biomedical domains.

\paragraph{Summarization-Based Approach: Instruction-Tuning on Smaller Models\\}

Incorporating summarization improved the performance of generative models. By reducing contextual noise and narrowing the focus to task-relevant information, summarization made it easier for models to extract relations. Among generative approaches, Llama 3.2-3B-Instruct achieved the highest F1 score of 59.7\%, representing a substantial improvement over its zero-shot performance of 3.78\%. With this method, the model produced more aligned, concise, and consistent outputs; reduced hallucinations; and demonstrated a greater ability to map its responses to predefined relation classes.

In contrast, the fine-tuned Mistral 7B-Instruct model continued to exhibit several issues encountered in zero-shot setting. Although fine-tuning provided a moderate improvement in classification accuracy, hallucinations and variability in output formatting persisted. For instance, the model generated semantically similar but structurally inconsistent relation types such as \textit{experience}, \textit{experiences}, and \textit{experiencer}. It also introduced relations not present in the training data, suggesting a tendency to overgeneralize beyond the expected label set. Instruction-tuning Llama3 8B followed the same trend. The inconsistencies in the hallucinations and the output format are better than the ones of the zero-shot approach, but are still persistent regardless of the instruction tuning.

%Notably, Mistral also produced extended text spans instead of relation labels alone. These outputs often included justifications or paraphrased versions of the relation, which, while potentially informative, are unsuitable for structured RE where strict label conformity is required.

The performance gap between instruction-tuned models can be partly explained by their number of trainable parameters. Llama 3.2-3B (24M) and Llama3 8B (50M) adapted well to the small MicrobioRel dataset, while Mistral 7B (95M) showed signs of overfitting and hallucination. Larger models typically require more data to generalize, and limited supervision can lead to unstable outputs, as supported by prior work \cite{gekhman2024does}.

%The difference between the two instruction-tuned models is partly explained by the number of trainable parameters during LoRA fine-tuning. Llama 3.2-3B has approximately 24M trainable parameters, while Mistral 7B requires nearly 95M. With a relatively small dataset like MicrobioRel, this disparity plays a significant role. Larger models with more trainable parameters generally require larger and more diverse datasets to generalize well and avoid overfitting. Mistral's larger capacity may have led to suboptimal adaptation, reflected in the inconsistency and hallucinations observed. These observations are in line with prior findings~\cite{gekhman2024does}, which associate hallucinations with the integration of novel information during fine-tuning.
%In contrast, Llama’s smaller size and fewer trainable parameters made it more suitable for adaptation in a low-resource setting. Its outputs demonstrate higher consistency and adherence to the target label space, making it a more reliable candidate for RE tasks under resource constraints.

%\secrev{On the generative approach, zero-shot RE using long prompt in the context of document-level RE and multiple relation types is not effective. Instead, reducing textual noise through summarization and encoding task-specific constraints through instruction-tuning, even when working with small, specialized datasets, proves to be a better approach for aligning generative models with structured extraction objectives.}

\paragraph{Regular Fine-tuned Models\\}

%\secrev{
BERT-based biomedical models such as PubMedBERT, BioBERT, SciBERT, and BioLinkBERT outperform all other tested models, with PubMedBERT achieving the highest F1 score. This highlights the effectiveness of domain-specific pretraining and regular fine-tuning for RE in biomedical texts. Despite the rise of instruction-tuned and generative models, specialized encoders remain highly effective, especially when task formulation is well aligned with traditional classification.%} 

Overall, our findings show that while zero-shot RE with generative models struggles in a specialized multi-class biomedical RE, combining instruction tuning with summarization leads to marked improvements, particularly for smaller, more adaptable models. Instruction tuning helps enforce task constraints, and summarization reduces contextual noise, making this strategy especially promising in low-resource settings.
Although fine-tuned encoder models like PubMedBERT still achieve the best overall performance, generative models can be presented as an alternative when annotation resources are limited. In future work, it would be valuable to explore the threshold at which datasets become too small for traditional fine-tuning to remain effective, opening the door for lightweight, tuned generative models to be better fitted for the task.

\subsection{Summary Quality Evaluation}

To assess the quality of the generated summaries, we evaluated their semantic similarity to the original passages using two complementary metrics: cosine similarity and F1-BERTScore. 

%\commenaireSolen{On peut réduire la taille de la figure 2, si besoin}

\begin{figure*}[!hbt]
\begin{center}
\includegraphics[width=0.45\textwidth,height=0.3\textheight]{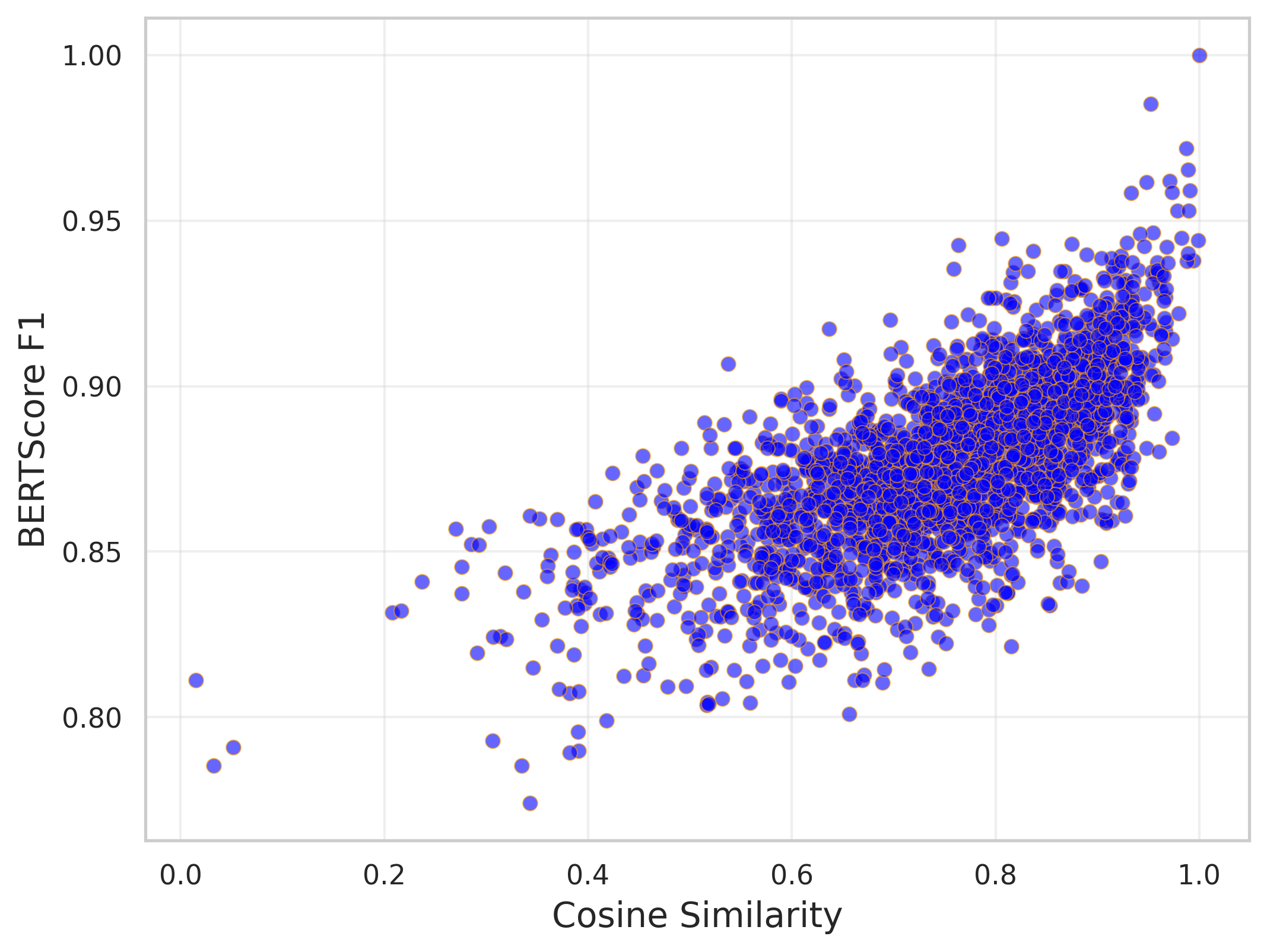}
\caption{Relationship between Cosine Similarity and F1-BERTScore, for the evaluated textual pairs.}% The plot shows a positive correlation, with higher Cosine Similarity generally corresponding to higher F1-BERTScore values, highlighting the alignment between these metrics in capturing semantic relevance}
\label{scatterp}
\end{center}
\end{figure*}

As shown in Figure~\ref{scatterp}, there is a moderate to strong positive Pearson correlation between these metrics, indicating that both capture semantic alignment reliably. Interestingly, F1-BERTScore remains relatively high (0.80–0.90) even when cosine similarity is low (< 0.5), highlighting its robustness in detecting semantic overlap beyond surface-level similarity. When cosine similarity is high (approaching 1.0), both metrics converge, suggesting strong alignment between summary and original text.

Additionally, correlation analysis revealed that longer original passages tend to slightly reduce similarity scores, which is likely due to increased textual noise, while longer summaries are weakly associated with higher similarity. Overall, this evaluation supports that the generated summaries preserve the core semantic content needed for downstream RE.

\subsection{Error Analysis}

Given that our summarization-based method is a two-step pipeline, it is susceptible to error propagation. To better understand the origin of performance limitations, we conduct an error analysis aimed at distinguishing errors coming from the summarization step from those introduced during the relation classification phase. We manually inspect a subset of 30 examples from the test set where the predictions diverge from the gold annotations. These cases are categorized into three distinct error types: errors from the summaries, where the summary does not contain the relevant interaction (Class 1), counting 6 elements, errors originating from the RE model (Class 2), counting 13 elements, and predictions that, although different from the annotation, are valid and semantically plausible interpretations of the input (Class 3), counting 11 elements.
%\begin{itemize}
%    \item Errors from the summaries, where the summary does not contain the relevant interaction (Class 1); 
%    \item Errors originating from the RE model (Class 2);
%    \item Predictions that, although different from the annotation, are valid and semantically plausible interpretations of the input (Class 3).
%\end{itemize}
\begin{comment}
\begin{table}[h!bt]
    \centering
    \begin{tabular}{|l|c|}
        \hline
        \textbf{Error Class} & \textbf{Count} \\
        \hline
        Class 1 & 6 \\
        Class 2 & 13 \\
        Class 3 & 11 \\
        \hline
    \end{tabular}
    \caption{Error distribution of manual analysis on 30 examples from the test set}
    \label{error_distribution}
\end{table}
\end{comment}

%\subsubsection*{Class 1: Errors from the Summaries}
\paragraph{Class 1 Errors} The fewest errors were attributed to the summarization step (Class 1), suggesting that the summaries generally preserved the key interactions necessary for RE. This indicates that the summarization phase was effective in retaining relevant information. An illustrative example of this type of errors is presented below:

\begin{quote}
\textbf{Original text:} "Nevertheless, the role of FXR in diabetes is controversial. Although study has reported that upregulating FXR in diabetic mice can significantly improve hypercholesterolemia, other study found that inhibition of intestinal FXR can reduce hepatic gluconeogenesis, promote the secretion of GLP-1, and reduce body weight. Therefore, whether upregulating or inhibiting FXR signaling is an innovative approach in the control of blood glucose in patients with T2D needs further research."
\textbf{Generated Summary:} "Unfortunately, there is no relation mentioned between FXR (Farnesoid X Receptor) and blood glucose in the given text. The text discusses various topics such as depression, hypertension, gut microbiota, and diabetes, but it does not mention a direct relationship between FXR and blood glucose. "

\textbf{Extracted Relation:} None
\textbf{Annotated Relation:} Affects 
\end{quote}

\noindent In this example, the relevant entities are \textit{FXR} and \textit{Blood Glucose}. While the original passage discusses the potential influence of FXR signaling on blood sugar regulation in type 2 diabetes (T2D), the summary fails to catch this relation. This can be explained by the speculative nature of the original sentence, which could lead the summary to focus on general context rather than nuanced interactions. The extracted relation, therefore, remains correct relative to the generated summary.

%\subsubsection*{Class 2: Errors from the RE Model}
\paragraph{Class 2 Errors} Class 2 errors are mainly common when the expected relation is linked to a quantity or a property. The relation \textit{Increase} is particularly present in the examples belonging to this class, because the relation does not refer directly to the entity itself, but to an attribute, resulting in the prediction of a different relation.
\begin{quote}
\textbf{Original text:} "Patients suffering from PD have different bacterial flora compared with healthy controls. Microbiota have an impact on disease progression, which has been established in the fecal-transplantation experiments. Administration of the microbiota of patients with Parkinson's disease to mice showed the development of neuroinflammatory processes and motor deficits."
\textbf{Generated Summary:} "**Relation:** The bacterial flora in patients suffering from PD is different from that of healthy controls.**Effect:** The altered microbiota can impact the progression of PD, specifically leading to neuroinflammatory processes and **motor deficits**."

\textbf{Extracted Relation:} Associated with
\textbf{Annotated Relation:} Increase
\end{quote}

\noindent In the example above, the relation to capture is between \textit{Parkison's disease} and \textit{motor deficits}. The reflected relation in the generated summary focuses on the entire context of the text, which includes bacterial flora of PD patients. This mismatch arises because the model predicts a broad association rather than the specific directional change implied by the annotated relation. The increase in motor deficits is indirectly expressed and mediated by microbiota, making it harder for the model to link it causally to Parkinson’s disease.
%Solen : ajouter ici un exemple réel d'erreur, comme pour les 2 autres cas d'erreur

%\subsubsection*{Class 3: Errors Compared to the Annotated Relations}
\paragraph{Class 3 Errors} For Class 3 errors, the model’s predictions are in fact reasonable, despite being different from the annotated relations. These examples highlight the divergence between human annotation and the model’s interpretation, revealing that both can arrive at different yet reasonable understandings of the same passage. Many of these examples involve the automatically assigned None class, suggesting that some relations captured by the model may have been missed during annotation. 

\begin{quote}
\textbf{Original text:} "Probiotics Mixture Increases Butyrate, and Subsequently Rescues the Nigral Dopaminergic Neurons from MPTP and Rotenone-Induced Neurotoxicity."
\textbf{Generated Summary:} "The relation between MPTP (1-methyl-4-phenyl-1,2,5,6-tetrahydropyrine) and Nigral Dopaminergic Neurons is that:MPTP-induced neurotoxicity can damage nigral dopaminergic neurons, but a probiotics mixture can increase butyrate levels, which in turn rescues these neurons from MPTP-induced neurotoxicity."

\textbf{Extracted Relation:} Worsen
\textbf{Annotated Relation:} Affects 
\end{quote}

In summary, most errors originate from the RE step rather than the summarization one, which generally preserves the semantic information of interest. The analysis highlights the challenge of balancing contextual understanding with precise semantics, especially in subtle or ambiguous cases. It also points out the limitations of automatically assigning "no relation" labels, and the need for refinement in handling nuanced interactions.

%%================================================================
\section{Conclusion}

This on-going work investigated the use of LLMs for biomedical RE, introducing summarization as an effective intermediate step. The approach helped reduce textual noise and better align model outputs with structured task requirements, even in low-resource settings. While promising, the study also highlighted persistent challenges such as hallucinations, inconsistent outputs, and difficulty capturing subtle context-dependent relations. These findings point to the need for further refinement in generative RE pipelines, particularly in balancing semantic precision with contextual understanding, specifically in low resource specialized context.

\section{Limitations}
This study focuses exclusively on the MicrobioRel dataset, as our primary interest lies in the analysis of interactions within the gut microbiome. At present, MicrobioRel is one of the few available manually annotated resources having a multi-class setting (with more than ten relation types) and multiple entity types. However, we acknowledge the importance of expanding our evaluation to additional datasets in future work, even if they differ in scope as they could provide valuable perspectives and robustness checks for our approach.
Moreover, we do not compare our method directly with other generative RE approaches, as most of them either address simpler binary tasks or evaluate performance in an end-to-end way by including named entity recognition, making direct comparisons difficult. Adapting our pipeline to these settings would be a promising direction for future extensions.

Another limitation of our study is that we did not generate multiple summaries per passage or apply any selection strategy. Instead, we relied on a single generated summary, selected based on its semantic similarity to the original text. Future work could explore generating multiple summaries and selecting the most relevant one using task-specific quality criteria.

Finally, while our summary-based method shows encouraging results, traditional BERT-based RE models continue to outperform it in terms of raw extraction accuracy. This suggests that further research is needed to enhance the effectiveness of generative models, particularly if they are to offer a computational or practical advantage over current methods.

%parler du fait qu'il faut trouver un ratio données/modèles à utiliser (si on avait moins de données, il se peut que ce ne soit pas suffisant pour fine tuner un BERT classique, donc il vaut mieux tester un LLM.
%direct approach with summaries ?
%%================================================================
%\section*{Remerciements (pas de numéro)}

%Paragraphe facultatif, ajouté seulement dans la version finale (pas lors de la soumission).

%%================================================================
%% Note : si l'on préfère éviter de factoriser les crossrefs :
%% bibtex -min-crossrefs 99 taln-exemple
%%================================================================
\bibliographystyle{coria-taln2025}
\bibliography{biblio}
%\nocite{TALN2015,LaigneletRioult09,LanglaisPatry07,SeretanWehrli07}
\appendix
\section{MicrobioRel: Relations Statistics}
\label{dataset}

\begin{table}[!ht]
\centering
\begin{tabular}{lc|lc}
\hline
\textbf{Relation type} & \textbf{Count} & \textbf{Relation type} & \textbf{Count} \\
\hline
Associated\_with & 210 & Possible & 56 \\
Part\_of & 204 & Worsen & 51 \\
Affects & 202 & Marker-Mechanism & 41 \\
Increase & 185 & Prevents & 39 \\
Location\_of & 146 & Physically\_related\_to & 31 \\
Causes & 144 & Negative\_correlation & 23 \\
Experiences & 132 & Treats & 17 \\
Decrease & 121 & Complicates & 10 \\
Start & 118 & Presence & 9 \\
Improve & 102 & Reveals & 7 \\
Interacts\_with & 86 & Stop & 6 \\
\hline
\end{tabular}
\caption{Relation label counts in the MicrobioRel corpus}
\label{tab:relcount}
\end{table}

\section{MicrobioRel Example}
\label{example}
\begin{quote}
\textbf{Text:} "During PD development, microbiota changes can be accompanied by reduced concentrations of branched-chain amino acids (BCAAs) and aromatic amino-acids in comparison with the healthy control group."

\textbf{Entity 1:} PD \\
\textbf{Entity 2:} branched-chain amino acids \\
\textbf{Annotated Relation:} Decrease
\end{quote}

%%================================================================
\end{document}